\documentclass[runningheads]{llncs}
\usepackage{graphicx}
\usepackage{cite}
\usepackage[utf8]{inputenc}
\usepackage{subfig}

\begin{document}
\title{Measuring Unfairness through Game-Theoretic Interpretability\thanks{The first author has been supported by Magazine Luiza. The second author has been partially supported by CNPq grant 312180/2018-7.}}
%
%
\author{Juliana Cesaro \orcidID{0000-0003-2091-7994}
\and \\
Fabio G. Cozman \orcidID{0000-0003-4077-4935}}
\authorrunning{Cesaro and Cozman}
%
\institute{Escola Polit\'ecnica da Universidade de São Paulo, USP, Brazil 
}
%
\maketitle              
\begin{abstract}
One often finds in the
literature connections between measures of fairness and
measures of feature importance employed to interpret trained classifiers.
However, there seems to be no study that compares fairness measures
and feature importance measures. 
In this paper we propose ways to evaluate and compare such
measures. We focus in particular on SHAP, a game-theoretic measure
of feature importance; we present results for a number of
unfairness-prone datasets.
\end{abstract}

\section{Introduction}

Machine learning algorithms have been used in a range of applications, from decisions about bank loans to criminal sentencing. Due to  concerns about algorithmic fairness~\cite{compas, amazon_tool}, several metrics have been created to detect bias injustice across  groups \cite{equal_opportunity, disparate_impact} or
individuals \cite{individual_fairness_measure, temporal_individual_fairness} or both \cite{inequality_indice}.
It has been suggested that measures of feature
importance can identify failure of fairness \cite{feature_importance_fairness, feature_importance_fairness2, feature_importance_fairness3} as 
feature importance can indicate that a
feature has a larger effect that it should have  \cite{shap_tree, characterization_feature_importance,shapley_sampling_values, lime}.

There seems to be no study that verifies whether feature importance
measures are indeed useful in assessing fairness. Moreover, no
study about the (supposed) connection between the two kinds of measures seems
to be available. This paper proposes a simple scheme to evaluate the relationship between feature importance and fairness measures, by comparing measures on a dataset with and without bias removal technique.

The contributions of this paper are: (i) a framework  to evaluate the merits of feature importance in assessing fairness, based on comparing the variation of results with and without application of  reweighing; and (ii) a study across four standard datasets, where the results obtained through feature importance are compared with  fairness measures.
We focused on reweighing techniques to remove bias \cite{reweighing}
and SHAP to measure feature importance \cite{shap}. 

In the next section we review various metrics concerning
fairness and interpretability; we introduce a few twists
to emphasize their connection. Later we describe our proposals
and experiments.

\section{Background}

In this section we summarize definitions of fairness, techniques
for bias removal, and tools to measure feature importance
that are related to interpretability. 

\subsection{Defining fairness}

Definitions of fairness can be divided into two major categories: group fairness and individual fairness. These definitions quantify the relationship between an ``unprivileged"  and a ``privileged" group. 

In this paper we assume that there is a unique {\em sensitive} feature $A$ that differentiates the privileged group
from the unprivileged one. And we assume that that value zero for this feature signals the  unprivileged group, while value  one indicates the privileged group. The target output has values $\{0,1\}$, where $1$ is the desirable class, such as good credit score, and $0$ is the undesirable class. 

\textbf{Group fairness} is obtained when the privileged and the unprivileged groups are treated the same. One possible way to quantify group fairness is to use  {\em disparate impact}  \cite{disparate_impact}:
\begin{equation}\label{disparate_impact}
disp\_impact = \ {P(\hat{Y}=1|A=0)}/{P(\hat{Y}=1|A=1)},
\end{equation}
where $\hat{Y}$ is the predicted outcome and $A$ is the sensitive feature. Equation (\ref{disparate_impact}) must be close to one to indicate fairness; other values indicate unequal treatment through feature $A$.

Another  measure of group fairness is based on predicted and actual outcomes as captured by {\em equality of opportunity}~\cite{equal_opportunity}:
\begin{equation}\label{equal_opportunity}
equal\_opport = P(\hat{Y}=1|A=0, Y=1)- P(\hat{Y}=1|A=1, Y=1),
\end{equation}
where $Y$ is the actual outcome. Expression (\ref{equal_opportunity}) should be close to zero; other values indicate unequal treatment.

Approaches that aim at equalizing relationships between groups may increase unfairness amongst individuals. Consider for instance a job application setting:  to equalize relationship between groups one may select less qualified candidates from the unprivileged group. 
Individual-level fairness then makes sense.

\textbf{Individual fairness} requires   similar individuals to receive similar classification outcomes. For instance,
{\em consistency} compares a model prediction of an instance $x$ to its k-nearest neighbors, $kNN(x)$ \cite{consistency}:
\begin{equation}\label{consistency}
consistency =  1 - \frac{1}{N} \sum_{n=1}^N \left| \hat{y}_n -  \frac{1}{k} \sum_{j \epsilon kNN(x^\prime_n)} \hat{y}_j \right|.
\end{equation}
Note that we here introduced a small change to the original formulation of consistency: instead of calculating   $kNN$ for the input $x$, we use $x^\prime$, where the latter refers to the input $x$ with  the removal of the sensitive feature. 
Expression~(\ref{consistency}) must be close to one to indicate fairness. 
 
 In this paper we employ the three definitions presented above in our experiments: disparate impact (\ref{disparate_impact}), equality of opportunity (\ref{equal_opportunity}) and consistency (\ref{consistency}). 
 
\subsection{Removing bias}

Techniques that attempt to remove bias from a model can be
divided into three categories: ones that  preprocess  data before a classifier is trained  \cite{pre_processing, disparate_impact, consistency}; ones that operate inprocessing, where the model is optimized at training time \cite{adult_feature}; and
ones that explores postprocessing of the model prediction \cite{equal_opportunity}. In this paper we adopt a preprocessing methodology called {\bf reweighing}  \cite{reweighing}  that aims at improving group fairness,
as it is a well-known technique that requires no hyperparameters
(thus allowing us to avoid lengthy digressions into parameter
tuning). In addition, reweighting does not change the features as other methods do \cite{disparate_impact, consistency}.

 Reweighing assigns weights to the points in the training dataset to reduce bias. Every random unlabeled data object $X$ is assigned a weight:
\begin{equation}\label{reweighing}
W(X) =  \frac{P_{exp} (A=X(A) \wedge Y=X(Y))}{P_{obs} (A=X(A) \wedge Y=X(Y))},
\end{equation}
where $P_{obs}$ is the observed probability, $P_{exp}$ is the expected probability and $A$ is the sensitive feature.
Lower weights are assigned to instances that the privileged class  favors. This approach is restricted to a single binary sensitive attribute and a binary classification problem. 

\subsection{Feature importance}

We can divide techniques
that explain the behavior of machine learning algorithms in two main groups: global approaches that aim at understanding the behavior of the model as a whole \cite{feature_importance_rf} and  local approaches that interpret individual predictions \cite{shap, shap_tree, lime}. In this paper we choose a local approach called SHAP, which has the advantage of ensuring three important properties: local accuracy, missingness and consistency. We use a local approach because if the methodology were used in practice, it is important that it be able to provide justification for a certain generated result, which would allow to assess whether the prediction was fair.

 SHAP (SHapley Additive exPlanations) \cite{shap} produces a local explanation for each prediction of a given classifier. Using insights from game theory, SHAP can explain prediction of any machine learning model and unifies concepts of several previous methods \cite{shapley_regression_values, shapley_sampling_values, lime, DeepLIFT, QII_Datta, layer_wise_relevance_propagation, tree_interpreter}. 
 
 SHAP approximates locally the function to be explained, which we call $f$, by a linear function
 $g$ such that 
 \[ \label{shap_accuracy}
f(x) = g(x^{\prime}) = \phi_0 + \sum_{i=0}^{M} \phi_{i}x^\prime_i,
\]
where $x^\prime$ is again the modified input  and each weight $\phi_i$ is called a SHAP value, given by
\begin{equation}\label{shap_value}
\phi_i = \sum_{S \subseteq N\backslash\{i\}} \frac{|S|!(M - |S| - 1)!}{M!}[f_x(S \cup \{i\}) - f_x(S)],
\end{equation}
where $S$ is the set of non-zero entries in $x^\prime$ and $N$ is the set of all input features. 

SHAP values satisfy a few properties.
{\em Local accuracy} requires the result of the explanation model $g$ for an input $x$ to be equal to the prediction of the model desired to explain $f$. 
{\em Missingness} requires   features missing in the input to be given no importance. Finally, recall that {\em consistency} states that if a change in the model occurs so that a feature has larger impact on the result, the importance of that feature should not decrease. 

In this paper we mostly focus on a unique sensitive feature  $A$ in the model, but with the graphical results provided by the SHAP framework one can understand influences that go far beyond that. For instance, one can investigate how the input of each feature impacts the output, analyze relations between variables, and verify which variables exert a greater influence on the model result.  

We selected two types of graphs that influences detected by SHAP:  dependence  plots and  summary plots \cite{shap_tree}.

Dependence plots represent the effect of a single feature in the model output. To represent this relation the plot shows in the x-axis the value of the feature and the y-axis shows the SHAP value of the same feature. SHAP dependence plots also let one visualize the effect of the feature with the strongest interaction (calculated by SHAP interaction values). These effects are shown by coloring from low (blue) to high (red) each dot in the graph with the value of an interacting feature. Examples of these graphs are shown later in Figure \ref{adult_dep_plot_lr}. 

Summary plot sorts features by  global impact on the model, calculated as
\begin{equation}\label{shap_global}
G_j = \frac{1}{N} \sum_{i=1}^N \mid	\phi_j^i \mid.
\end{equation}
Each dot in the graph represents the SHAP value of that feature.  Examples  are shown later in Figures \ref{adult_summary_plot_lr}. 

\section{Proposal}

In this section we propose techniques that will allow us to compare fairness measures and results obtained
through SHAP. It should be noted that simply computing SHAP values will not help us doing it: SHAP values
are computed by datapoint, whereas fairness measures capture the whole behavior of a classifier. Hence
the need for novel ideas as proposed here.


To evaluate fairness we resort, first, to global impact of each feature
(Expression (\ref{shap_global})) as we focus on the ranking of the feature in a list of features ordered
by descending global values. Besides looking at global impact, we also employ
the following measure:
\begin{equation}\label{diff_shap}
D_j = \frac{1}{N_k} \sum_{k = 1}^{N_k} \phi_j^k - \frac{1}{N_l} \sum_{l = 1}^{N_l} \phi_j^l,
\end{equation}
where $k$ represents unprivileged group and $l$ privileged group, and each $\phi$ is a SHAP value.
 A value of $D_j$ close to one indicates fairness, while a negative value   favors the privileged group and a positive value favors the unprivileged one.


\begin{figure}[t]
  \centering
  \includegraphics[width=10cm]{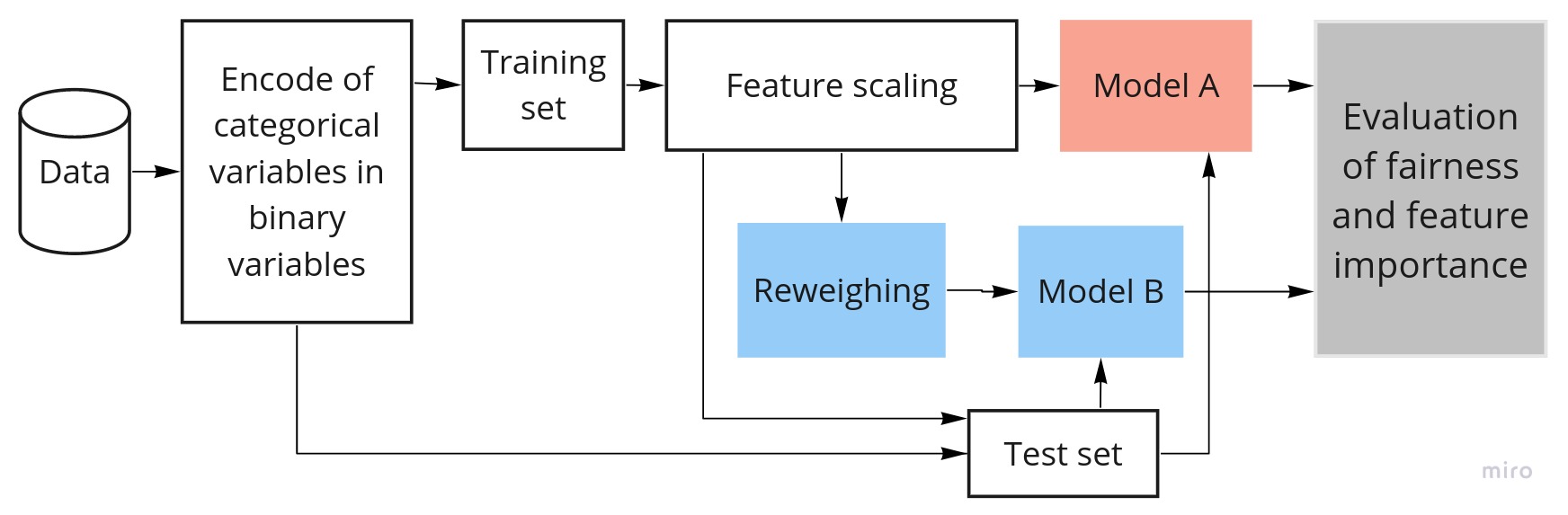}
  \caption{Fairness through feature importance: the model in red was trained with a fairness-sensitive dataset; the model in blue had bias removed.}
  \label{fluxo_processo}
\end{figure}

 Figure \ref{fluxo_processo} summarizes the steps in assessing  fairness through feature importance. An initial step in the workflow is the encoding of categorical variables in the dataset, followed by data split into $80\% - 20\%$ train-test sets and a standardization of features. 

Then there are two possible paths: the red box indicates training  directly  with the model, and the blue box indicates an additional step of de-biasing before model training (through reweighting). Finally, results obtained from feature importance and fairness measures are compared.

More precisely, we compare the results by evaluating how fairness and feature importance measures vary as bias varies.
When bias is present, we expect discrimination to appear in fairness measures (disparate impact and consistency smaller than one, and equality of opportunity smaller than zero), while we expect feature importance measures to display a negative SHAP value difference between privileged and unprivileged groups. 

Note that reweighting focuses on group fairness; consequently, we can expect three scenarios concerning group fairness, as we now analyze:
\begin{itemize}
\item  \textbf{Equality between groups}: this scenario is characterized by disparate impact close to one and equality of opportunity to zero. We hypothesize that privileged and unprivileged groups get similar importance, which should be reflected in their mean SHAP values getting closer and in some reduction in feature importance with a global SHAP value close to zero. 
\item \textbf{Favoring the privileged group}: this scenario is characterized by increase of disparate impact and equality of opportunity. However, disparate impact would remain smaller than one and equality of opportunity negative. We hypothesize a decrease in SHAP value difference, but the value would remain negative. We also hypothesize a decrease in importance of the sensitive feature. 
\item \textbf{Favoring the unprivileged group}: this scenario is characterized by inversion of importance between groups, which would be perceived with disparate impact result greater than one or equality of opportunity positive. We hypothesize the SHAP values difference between groups to be positive. While we expect an increase in the feature importance if the discrimination between groups increases, which would be perceived for example with increase in module of equality of opportunity, and we hypothesize a decrease in feature importance if the discrimination decreases.
\end{itemize}

Clearly the hypotheses just described must be validated through empirical analyses. This is the goal
of the remainder of this paper. Before we proceed, a comment on individual fairness: as reweighting
does not focus on individual fairness, it is hard in principle to say how reweighting affects consistency
(later we show that the relationship between these techniques is significant and actually somewhat
surprising).

\section{Experiments}

To test our proposed scheme and the hypotheses outlined at the end of the previous section, 
we applied Logistic Regression, Random Forests and Gradient Boosting to four unfairness-prone datasets
(using the  scikit-learn library\footnote{http://scikit-learn.org}).
The study was limited to one binary sensitive attribute and a binary classification problem, due to limitations in reweighing and in some fairness measures. However, the methodology used to obtain feature importance could be applied to any classifier, and the sensitive variable could be of any type.

 All tests were done using the same hyperparameters. The AIF-360 \footnote{https://aif360.mybluemix.net} library was used to apply reweighing and to calculate disparate impact and equality of opportunity metrics. We use the $kNN$ implementation of sckit-learn to compute the consistency metric.
 
 All datasets and techniques are available in a github repository.\footnote{https://github.com/cesarojuliana/feature\_importance\_fairness}

\subsection{Datasets}

Four datasets often analyzed with respect to fairness were used: Adult, German, Default and COMPAS datasets. 
Adult, German and Default datasets were obtained from the UCI repository \cite{uci} and the COMPAS dataset from ProPublica \cite{compas}.
The Adult dataset \cite{adult} contains information from the 1994 census database. The objective is to predict whether income is larger 50K dollars per year. We consider gender as the sensitive attribute, following Ref.\ \cite{adult_feature}. The German dataset contains information about bank account holders, and the goal is to classify  each holder as good or bad credit risk. We use age as the sensitive attribute as in Ref.\ \cite{pre_processing}. The Default dataset \cite{default} contains information from credit card clients in Taiwan from April 2005 to September 2005. The objective is to predict default of their customers. We use the gender as the sensitive attribute as in \cite{convex_framework_fair_regression}. The COMPAS dataset \cite{compas} contains data from criminal defendants in Broward County, Florida, which objective is to predict recidivism over a two-year period. We use the same filter as in Ref.\ \cite{compas_analysis}. The sensitive attribute is race, being selected Caucasian as the privileged group.

We remove the variable {\em fnlwgt} from Adult dataset, because this variable does not aggregate information to the problem goal. In the Default dataset we excluded {\em id} variable for the same reason. For the COMPAS dataset we use only the following variables: $race$, $age$, $c\_charge\_degree$, $v\_score\_text$, $sex$, $priors\_count$, $days\_b\_screening\_arrest$, $v\_decile\_score$, $two\_year\_recid$, $is\_recid$.

\subsection{Results and discussion}

\begin{figure}
  \centering
  \includegraphics[width=1\linewidth, height=7cm]{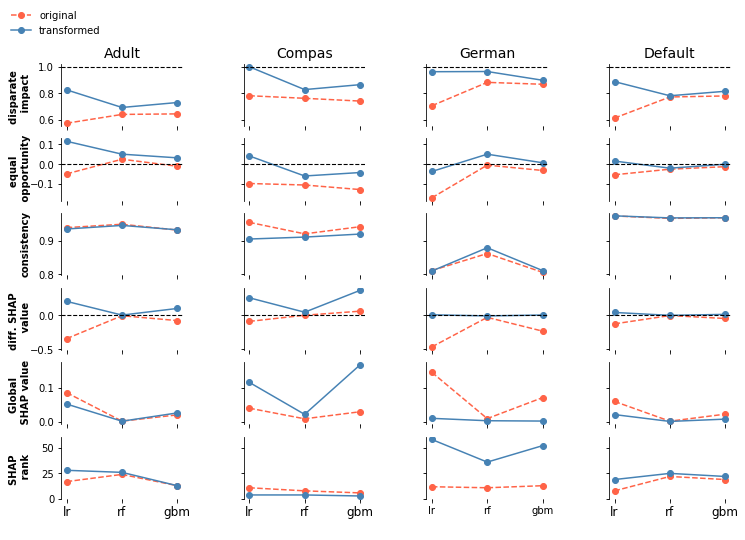}
  \caption{Evaluation of fairness on test set for four datasets: Adult, COMPAS, German and Default, for three types of models: Logistic Regression, Random Forest and Gradient Boosting. Fairness is evaluated according to several definitions so as to capture both group and individual fairness.}
  \label{resultados_experimento}
\end{figure}

\begin{figure}
  \centering
  \includegraphics[width=1\linewidth, height=7cm]{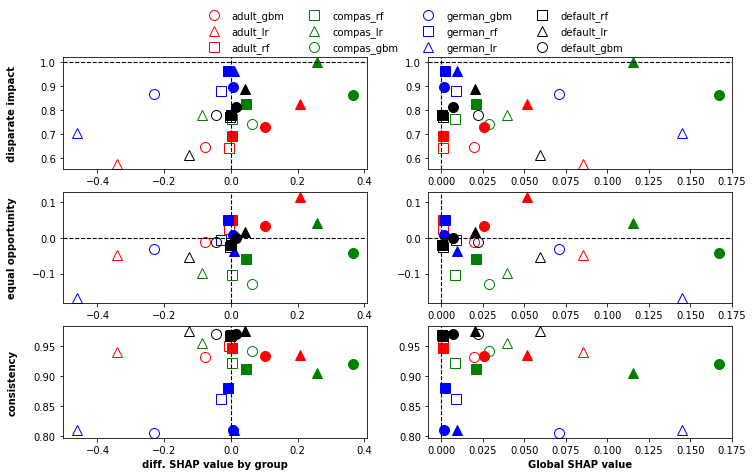}
  \caption{Comparison between fairness and feature importance measures, performed with four datasets (Adult, COMPAS, German and Default) and with three types of models (Logistic Regression, Random Forest and Gradient Boosting). The unfilled markers represents results without reweighing, and the filled markers results with reweighing.}
  \label{resultados_experimento_comparativo}
\end{figure}

In Figure \ref{resultados_experimento}, in the x-axis  $lr$ means Logistic Regression, $rf$ means Random Forest and $gbm$ means Gradient Boosting. The y-axis carries the names of fairness and feature importance measures. The red line shows   results without reweighing and the blue line shows results with reweighing. The results were separated in four columns according to the used dataset. In Figure \ref{resultados_experimento_comparativo} we can see the relation between variation in fairness measures with feature importance measures in the y-axis and the x-axis respectively. The unfilled markers represents results without reweighing, and the filled markers results with reweighing. The markers in red show results for Adult dataset, in green for COMPAS dataset, in blue for German dataset, and in black for Default dataset. Legends indicate both the dataset and abbreviation of the used model.

In this section we will classify the results according to the three scenarios described previously: {\em equality between groups}, {\em favoring the privileged group} and {\em favoring the unprivileged group}. This classification is made based on the fairness results, and we hypothesized about what would be the feature importance result. We will compare whether the assumptions made actually occurred.

From the results we see that when reweighing is applied, disparate impact, equality of opportunity and difference in SHAP value between groups had a variation greater than or equal to zero. Furthermore, in several cases equality of opportunity and difference in SHAP value changed from negative to positive value; there was no case where disparate impact changes to a value greater than one with reweighing. With exception of the COMPAS results, in all other cases there was an decrease in feature importance with reweighing.

In the COMPAS results we perceived the following peculiarity  with reweighing: increased in consistency and in feature importance. In COMPAS we also note the scenario of favoring the privileged group with Random Forest and with Gradient Boosting. However,   this case had the unexpected result of increasing in feature importance probably due to variation in consistency.

The scenario of equality between groups can be seen in the  Default datatset and German dataset with Gradient Boosting. In all cases we see that equality of opportunity was very close to zero, but disparate impact was not close to one. In this situation, feature importance measures behaved as expected, SHAP values difference and feature importance approached   zero.

We find the scenario of favoring the privileged group in Adult dataset with Logistic Regression and Random Forest, COMPAS with Logistic Regression and German with Random Forest. Only with Adult dataset there is no variation in consistency, and in this situation we see the expected scenario where equality of opportunity ranged from a negative to a positive value.


Results demonstrate two important facts. First, there is a direct relation between SHAP value difference and equality of opportunity, which is much more significant than the relationship with disparate impact. This is most evident from Figure \ref{resultados_experimento_comparativo}. Second, the relation of feature importance is inverse with consistency. In the results where there was a decrease in consistency, we note that the impact  of consistency dominates feature importance (rather than the effect of increase in equality of opportunity). 

Thus we reach our main conclusion in this empirical study: feature importance measures are connected {\em both}
with consistency and equality of opportunity. Consequently we see that feature importance measures
do quantify {\em both} group and individual fairness. 

\begin{figure}[t]
\subfloat{(a)}{\includegraphics[width=5.5cm, height=4cm]{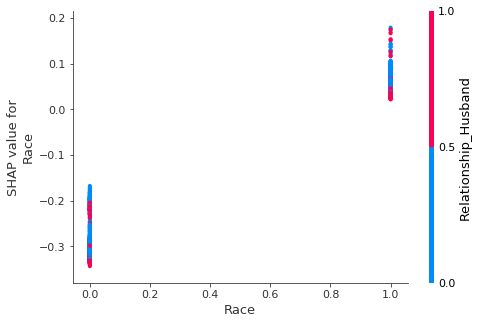}}
\subfloat{(b)}{\includegraphics[width=5.5cm, height=4cm]{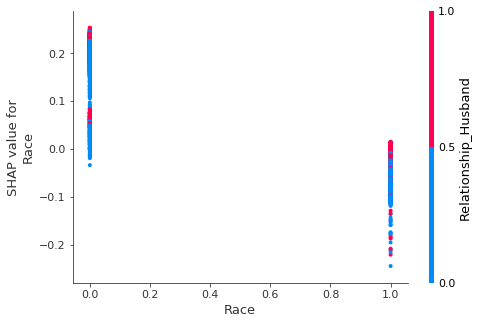}}
\hfill
\caption{SHAP dependence plot of a Logistic Regression trained with Adult dataset. The sensitive feature is race, where value one refers to white and value zero to non white. In y-axis is the SHAP value attributed to race. In (a) the model was trained with the original data, and we can see that it associated higher SHAP values with the white group, indicating discrimination in the dataset. In (b) reweighing was applied to reduce unfairness caused by  race; we note that the relationship was reversed and non white became the group favored by the model according to SHAP values.}
\label{adult_dep_plot_lr}
\end{figure}

\begin{figure}[h]
\centering
\subfloat{(a)}{\includegraphics[width=5.5cm]{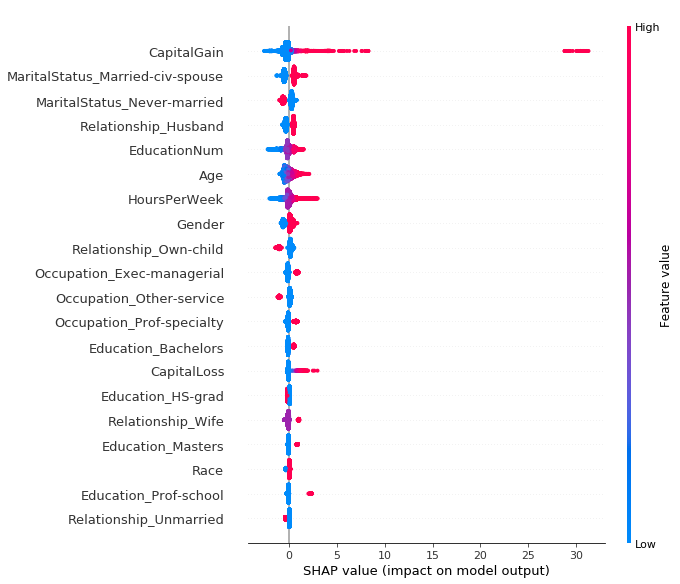}}
\subfloat{(b)}{\includegraphics[width=5.5cm]{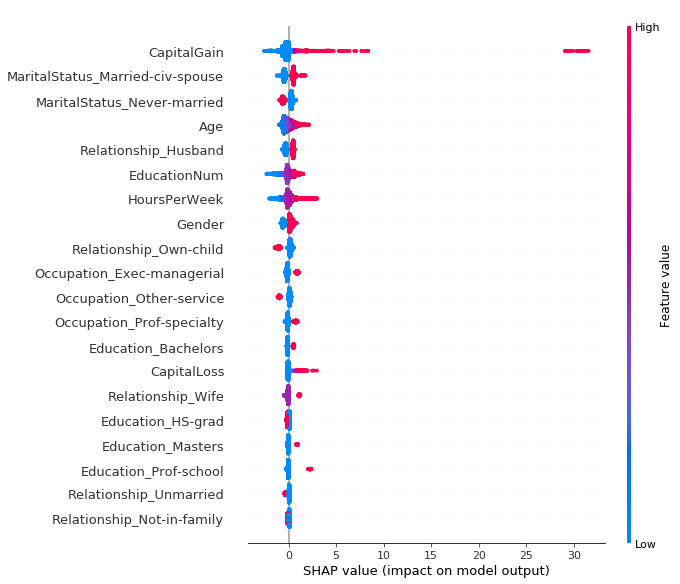}}
\caption{SHAP summary plots of Logistic Regression trained with Adult dataset not applying (a) and applying (b) reweighing so as to reduce unfairness caused by race. We can see that without reweighing the rank position of race was 17th. When reweighing is applied the position decreases, and race is no longer among the twenty most important features.}
\label{adult_summary_plot_lr}
\end{figure}



In the remainder of this section we present some additional remarks on the graphs that are provided
by graphs build with SHAP values and on the insights that one may get from them.
Basically, these graphs allow us to note implicit relationships between variables. 
Furthermore, they display the overall effect of {\em any} variable in the model by varying its input. 

For example, Figures \ref{adult_dep_plot_lr} and \ref{adult_summary_plot_lr}
show results obtained with Logistic Regression in the Adult dataset, with and without reweighing. Figure \ref{adult_dep_plot_lr} 
shows dependence plots, and Figure \ref{adult_summary_plot_lr} 
shows summary plots.

In the Adult dataset, race is the sensitive feature, and it was assigned value zero  for the unprivileged group and value one for the privileged group. 
Figure~\ref{adult_dep_plot_lr} depicts a partial dependency plot of Logistic Regression that displays unfairness between privileged and unprivileged group when reweighing is not applied, but the relation is inverted when reweighing is applied (greatly favoring the unprivileged group). 

In Figure \ref{adult_summary_plot_lr} 
we can see that, besides the decrease in rank position of the sensitive feature race, other variables changed in importance, such as the increase in $Age$ when reweighing is applied from the 6th to 4th position.

In short: it is difficult to get real insights on relationships amongst variables by examining SHAP values
and related graphs. A possible future research topic would be to extract such insights automatically.

\section{Conclusion}

We presented a framework that compares fairness definitions (group or individual) with results based on feature importance as quantified by SHAP. The basic idea is to examine how fairness definitions vary by changing the effect of the sensitive feature on the model (this was done here with   reweighing). Experiments show that feature importance measures can identify  group and individual fairness in the model. Certainly this is a preliminary effort that must be refined and extended in a variety of ways, but we feel that it is a valuable contribution due to the absence of similar analyses in the literature.

In particular, further work is needed to remove some important restrictions. We have focused on binary sensitive
features and two-class classification problems. Such restrictions must be lifted.
In future work we intend to study  other interpretability techniques that are based on feature importance. This would allow us to determine whether some methodologies work better for some definitions of fairness than others. Furthermore, we want to extend the tests to other techniques that remove bias besides reweighting. Another promising extension of this work would be
to evaluate the visualization techniques that must be used to present results. SHAP graphs speed up the perception of relationships between variables, but additional insights would be welcome.

%
%
%
\bibliographystyle{splncs04}
%
\bibliography{sample-bib}




\end{document}